\pdfoutput=1

\documentclass[11pt]{article}

\usepackage[preprint]{coling}

\usepackage{times}
\usepackage{latexsym}
\usepackage{times}
\usepackage{graphicx}
\usepackage{amssymb}
\usepackage{multirow}
\usepackage{booktabs,arydshln}
\usepackage{dcolumn}
\usepackage{array}
\usepackage{tabularx}
\usepackage{xcolor}
\usepackage{comment}
\usepackage{tablefootnote}
\usepackage{enumitem}
\usepackage{threeparttable}
\usepackage{cleveref}
\crefname{subsection}{subsection}{subsections}

\usepackage[T1]{fontenc}

\usepackage[utf8]{inputenc}

\usepackage{microtype}

\usepackage{inconsolata}

\usepackage{graphicx}

\title{CoCoP: Enhancing Text Classification with LLM \\through  Code Completion Prompt}

\author{Mohammad Mahdi Mohajeri \\
  University of Tehran\\
  \texttt{mehdimohajeri@ut.ac.ir} \\\And
  Mohammad Javad Dousti \\
  University of Tehran\\
  \texttt{mjdousti@ut.ac.ir} \\\And
  Majid Nili Ahmadabadi \\
  University of Tehran \\
  \texttt{mnili@ut.ac.ir} \\}

\begin{document}
\maketitle
\begin{abstract}
Text classification is a fundamental task in natural language processing (NLP), and large language models (LLMs) have demonstrated their capability to perform this task across various domains.
However, the performance of LLMs heavily depends on the quality of their input prompts.
Recent studies have also shown that LLMs exhibit remarkable results in code-related tasks.
To leverage the capabilities of LLMs in text classification, we propose the Code Completion Prompt (CoCoP) method, which transforms the text classification problem into a code completion task.
CoCoP significantly improves text classification performance across diverse datasets by utilizing LLMs' code-completion capability.
For instance, CoCoP enhances the accuracy of the SST2 dataset by more than 20\%.
Moreover, when CoCoP integrated with LLMs specifically designed for code-related tasks (code models), such as CodeLLaMA, this method demonstrates better or comparable performance to few-shot learning techniques while using only one-tenth of the model size.
The source code of our proposed method will be available to the public upon the acceptance of the paper.
\end{abstract}

\section{Introduction}
Large Language Models (LLMs) show commendable performance in both classification and generation tasks.
However, to enhance their performance, fine-tuning task-specific data is imperative.
The fine-tuning process requires collecting substantial, high-quality data pertinent to the domain task.
Also, fine-tuning is expensive and may not be suitable for all situations~\citep{kaddour2023challenges}.
On the other hand, in-context learning techniques, such as few-shot learning, offer more flexibility~\citep{GPT-2_few-shot_learner}.
Unlike fine-tuning, these techniques do not require extensive domain-specific data and work only with a few examples~\citep{GPT-2_few-shot_learner}.
It makes them accessible for anyone to use during inference.
It is important to emphasize that the effectiveness of these methods relies on the context and examples given as input.

LLMs are trained on massive datasets that include diverse input styles.
LIMA~\citep{zhou2023lima} claims that if a model can learn the style of interaction with the user, it can reach superior performance in alignment and effectively communicate its knowledge to the user.
This concept works in the reverse direction as well.
When users interact with LLMs using prompt structures familiar to LLMs, the model demonstrates better performance.
The efficacy of a prompt format for the model is contingent on its training data.
Code-related datasets exist on the internet and are among the most critical resources for the pre-training phase of LLMs.
For instance, LLaMA~\citep{llama} is pre-trained on a substantial 328 GB dataset sourced from GitHub, constituting 4.5\% of the overall model training data.
Additionally, Google has mentioned that code-related data are one of the most important parts of their pre-training data for the Gemini model~\citep{team2023gemini}.
Choosing code-related datasets highlights the importance of teaching LLMs about coding style and code creation.
Moreover, LLMs show good performance in code-related benchmarks.
For instance, both LLaMA and LLaMA2~\citep{llama2} have reported commendable results in code-related benchmarks.
The proficiency of LLMs in code-related tasks can be leveraged for tasks beyond coding.

To enhance LLMs performance in classification tasks, we introduce the \textit{Code Completion Prompt} (CoCoP) method in this paper.
Our suggested method uses LLMs' capabilities in code-related tasks, particularly in code completion.
Additionally, we employ the in-context learning in this method.
CoCoP creates incomplete-code which consists of a few demonstrations in code format.
Next, the LLM completes the code to determine the proper label for a user query.
Thus, CoCoP is founded on the strengths of LLMs' capabilities in code completion and in-context learning.

The effectiveness of our method is evaluated across various classification datasets using LLaMA2 and CodeLLaMA models~\citep{roziere2023codeLlama}, revealing superior performance compared to the few-shot learning method in classification tasks.
Also, our experiments indicate that LLMs designed for code-related tasks (code models) demonstrate superior performance with this method compared to other LLMs.
The source code of our proposed method will be available to the public upon the acceptance of the paper.

Our research contributions include:
\begin{itemize}
    \item {Introduced a novel text classification method using LLMs leveraging their code completion capability.}
    \item {Outperformed the traditional few-shot learning method in text classification with LLMs.}
    \item {Achieved comparable or superior performance using smaller code models like 7B and 13B model size compared to larger models such as the 70B model size across various text classification benchmarks through our approach.}
\end{itemize}

The rest of this paper is organized as follows.
\Cref{sec:method} describes our proposed method.
Next, \Cref{sec: results} shows results of CoCoP performance and impact of each part of it.
\Cref{sec: related work} describes related works.
Limitations and future work are discussed in \Cref{sec: lim&f }.
Finally, \Cref{sec : conclusion} concludes the paper.
\section{Method}
\label{sec:method}
\begin{figure*}[ht]
    \centering
    \includegraphics[trim=0cm 0cm 0cm 0cm, width=1\textwidth]{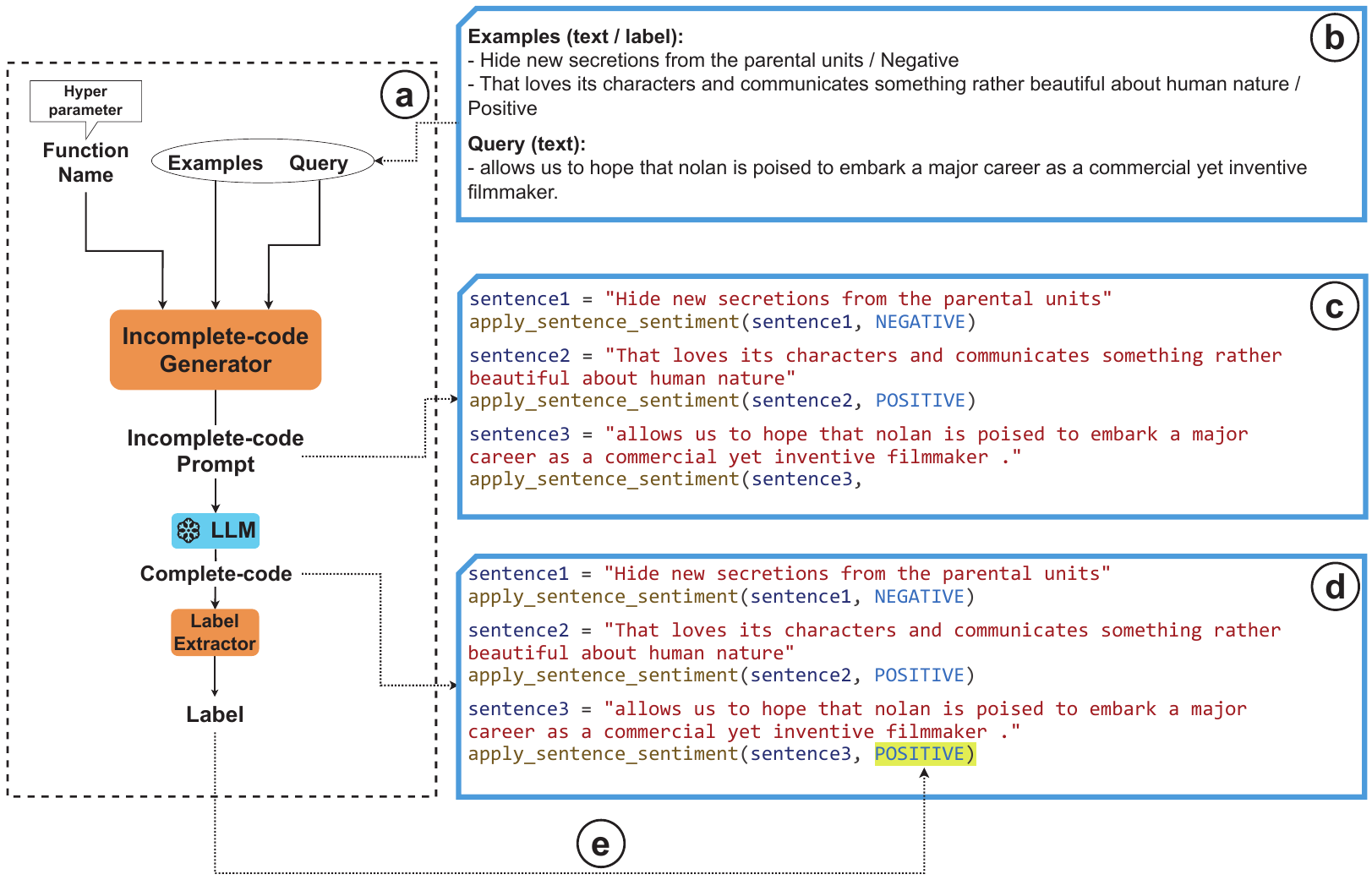}
	\caption{
    The figure illustrates the CoCoP method used for a sentiment analysis task.
    The process begins by providing examples from the training dataset and the query sentence to the Incomplete-code Generator module in plain text format (b).
    Then, this module generates incomplete code (c) and forwards it to the frozen LLM.
    The LLM completes the code and determines the label by code completion (d), passed to the Label Extractor module.
    Finally, the Label Extractor extracts the corresponding label (e) and outputs the final result.
    }
	\label{fig:outputs}
\end{figure*}

Our approach, called CoCoP, revolves around transforming the classification task into a code completion task.
We use it to improve LLMs's classification performance.
An overview of this method is illustrated in \Cref{fig:outputs}~(a).
The few-shot learning technique requires examples, so we employ examples for the CoCoP method.

These examples are passed to \textit{Incomplete-Code Generator} along with the query it should classify.
The Incomplete-Code Generator is the main module of our method.
This module accepts examples, query, and function name as inputs and produces incomplete-code as output.
The function name is a hyperparameter that the user should pass to this module.

Incomplete-Code Generator module performs two primary roles.
First, it converts examples and their labels into code, transforming each example into one or more variable assignments.
Then, utilizing a function call within the code, the label of each example is applied to it.
This function call uses a function name provided by the user.
Also, the name of the used function is crucial as it indicates the task process and guides the LLM in achieving better performance.
So, this method can be customized for various classification tasks by modifying the function name and incomplete-code structure.
Second, this module transforms the query into an incomplete function call.
This function call has the variable assigned with a query as a first argument, and the second argument, the label, is not passed to it. 
The LLM is expected to complete this function call with a correct label.

Given the LLM's proficiency in code completion, it is expected to complete the code following the provided examples, producing an output with a format similar to those examples.
We also expect the generated label to come from the labels provided in the examples.
Then, the LLM's output is passed to the Label Extractor module.
This module identifies a suitable label from the list of all labels at a particular position within the generated output.
If the label Extractor finds an acceptable label, it is chosen as the output label; otherwise, the output indicates that no proper label is generated.
For instance, \Cref{fig:outputs} shows the process for the sentiment analysis task.

The sentiment analysis task (SST2)~\citep{wang2019glue} includes more than 67,000 training data and about 1800 test data for labelling.
Three of them are randomly selected from the training data.
These examples and the query are passed to the Incomplete-Code Generator module (\Cref{fig:outputs}~(b)), which generates incomplete-code as shown in \Cref{fig:outputs}~(c).
In the incomplete-code, each example is a sentence transformed into a string variable assignment.
Additionally, this module utilizes an \texttt{apply\_sentence\_sentiment()} function to apply the label to each sentence.
The \texttt{apply\_sentence\_sentiment} as the function's name is a hyperparameter that the user passes.
This function takes two arguments: the variable corresponding to the example sentence as the first argument and the example's label as the second argument.
Similarly, this module processes the query but does not complete the \texttt{apply\_sentence\_sentiment()} function with the target label which does not exist.
Hence, the generated code is incomplete (see \Cref{fig:outputs}~(c)).
Furthermore, in cases where multiple sentences exist for each input, such as in Natural Language Inference (NLI) tasks, each sentence can be expressed in a separate variable assignment.
The simplicity and adaptability of this method make it suitable for a diverse set of classification tasks.
Finally, the output of the Incomplete-Code Generator is passed to the LLM.
LLM completes this code by adding a label as a second argument to the \texttt{apply\_sentence\_sentiment} function call, as illustrated in \Cref{fig:outputs}~(d).
Then, the output of LLM is passed to the Label Extractor.
This module extracts the assigned label for target input and presents it as the result of classification (see \Cref{fig:outputs}~(e)).
To validate the efficacy of this method, we assess its performance across various classification datasets.

\section{Results}
\label{sec: results}
\subsection{Evaluation setup}
We use LLaMA2-chat~\citep{llama2} and CodeLLaMA-Instruct~\citep{roziere2023codeLlama} as our baseline models. 
%
We use these models with temperature 0 to make our results reproducible.
Furthermore, we assess the efficacy of our method across various classification datasets.
We examine the GLUE~\citep{wang2019glue} dataset for text classification, selecting SST2, CoLA, and MRPC for binary classification tasks.
For the multi-class classification task, we use the SNLI~\citep{snli_emnlp2015} dataset.
To generate few-shot examples, we use instances from the training data of these datasets.
We use two examples from each class within SST2, CoLA, and MRPC datasets to generate few-shot examples in incomplete-code.
Also, for SNLI, we utilize a single example for each class to generate a few-shot examples of incomplete-code generation for this task.

\begin{table*}[ht]
    \centering
\begin{threeparttable}
\resizebox{0.95\textwidth}{!}{
\begin{tabular}{llllllllr}
\toprule
\multirow{2}{*}{\textbf{Method}} & \multirow{2}{*}{\textbf{Model}}& \multicolumn{4}{c}{\textbf{Dataset}} \\
\cmidrule{3-6}
   & & \textbf{SST2} & \textbf{CoLA} & \textbf{MRPC} & \textbf{SNLI}\\
\cmidrule{1-6} \multirow{1}{*}{ \textbf{SoTA} } &
--& 97.5$^\dagger$ & 88.6$^\ddagger$ & 93.7$^\mathsection$ & 93.1$^\mathparagraph$ \\
\cmidrule{1-6} \multirow{6}{*}{ \shortstack[l]{\textbf{Few}\\\textbf{shot}} } &
LLaMA2-7B-chat     & 67.7$\pm$4.1& 75.0$\pm$1.5& 60.4$\pm$0.4& 32.5$\pm$1.5      \\
 & LLaMA2-13B-chat      & 59.0$\pm$4.9& 59.9$\pm$8.7& 73.1$\pm$0.9& 33.1$\pm$0.3      \\
 & LLaMA2-70B-chat      & 82.9$\pm$0.7& 68.2$\pm$6.1& \underline{75.4}$\pm$2.8& \underline{63.0}$\pm$1.7      \\
 \cmidrule{2-6}
  & CodeLLaMA-7B-Instruct & 79.7$\pm$5.3 & 61.5$\pm$2.3 & 71.3$\pm$1.7 & 58.6$\pm$1.6 \\
 & CodeLLaMA-13B-Instruct & 49.7$\pm$6.2 & 56.0$\pm$7.8 & 69.5$\pm$2.3 & 50.6$\pm$1.9 \\
 & CodeLLaMA-34B-Instruct & 23.0$\pm$2.9 & 70.7$\pm$4.6 & 44.3$\pm$3.7 & 60.9$\pm$2.2 \\
\cmidrule{1-6}
\multirow{6}{*}{ \textbf{CoCoP} } & LLaMA2-7B-chat & 58.3$\pm$7.7& 72.7$\pm$1.1& 71.0$\pm1.1$& 38.3$\pm$2.8       \\
 & LLaMA2-13B-chat & 85.7$\pm$2.7& 70.3$\pm$1.5& 72.7$\pm$1.9& 45.7$\pm$3.7       \\
 & LLaMA2-70B-chat & 91.7$\pm$0.3& \underline{78.1}$\pm$2.3 & \underline{75.4}$\pm$2.0& \textbf{67.7}$\pm$2.1 \\
 \cmidrule{2-6}
 & CodeLLaMA-7B-Instruct & 92.2$\pm$1.0& 76.9$\pm$1.0& 72.7$\pm$0.7 & 59.5$\pm$1.6      \\
 & CodeLLaMA-13B-Instruct & \textbf{93.5}$\pm$0.5& 76.3$\pm$0.9& 72.3$\pm$0.7& 61.4$\pm$1.0      \\
 & CodeLLaMA-34B-Instruct & \underline{92.3}$\pm$0.8& \textbf{78.5}$\pm$1.2& \textbf{77.4}$\pm$0.8& 62.0$\pm$1.1      \\
\bottomrule
\end{tabular}
}
\begin{tablenotes}[para]
\centering
\footnotesize {
	\item[$\dagger$] \citep{raffel2020sst2_sota}
	\item[$\ddagger$] \citep{cherniavskii2022cola_sota}
    \item[$\mathsection$] \citep{jiang2020mrpc_sota}
    \item[$\mathparagraph$] \citep{wang2021snli_sota}
}
\end{tablenotes}
\end{threeparttable}

    \caption{Results of few-shot and CoCoP methods using different LLaMA2 and CodeLLaMA models, evaluated by accuracy across SST2, CoLA, MRPC, and SNLI datasets.
    Note that SoTA models have been trained on training data from selected datasets.
    Therefore, it is expected that these models show better performance than using general-purpose LLMs.
    The table shows that the combination of CoCoP with CodeLLaMA-7B in SST2 and CodeLLaMA-34B in CoLA and MRPC datasets performs better than LLaMA2-70B with both CoCoP and few-shot methods.
    Moreover, CoCoP with CodeLLaMA 7B and 13B also shows comparable or better results to LLaMA-70B.}
    \label{tab:Models size}
\end{table*}

\subsection{Method performance}
In this subsection, we evaluate the CoCoP method performance in different datasets and assess its capability to improve classification tasks.
We also present SoTA model results which have been trained on training data from selected datasets.
Therefore, it is expected that SoTA models show better results than using general-purpose LLMs.

\subsubsection{Few-shot learning vs. CoCoP}
\label{sec:few-shot learning vs. CoCoP}
We used two methods for classification: few-shot learning as a base method and CoCoP.
We used LLaMA2-chat with 7B, 13B, and 70B variations for evaluation.
\Cref{tab:Models size} shows the results of this experiment.
In LLaMA2-7B-chat, CoCoP only performed better in MRPC and SNLI, but in LLaMA2-13B-chat, it performed better in SST2, CoLA, and SNLI.
Also, LLaMA2-70B-chat showed better performance in SST2, CoLA, and SNLI and similar performance in MRPC.
These results show that CoCoP can improve performance in classification tasks.
Also, this impact becomes more remarkable when the model size increases because it improves performance in the larger models.

\subsubsection{Code models vs. base models}
\label{sec:coders vs. base models}
Considering our methodology, which emphasizes leveraging the capabilities of LLMs for code completion, we specifically concentrate on code models.
These models are fine-tuned on tasks related to coding and are trained with substantially larger amounts of code data~\citep{roziere2023codeLlama, luo2023wizardcoder}.
So, we used CodeLLaMA-Instruct as a code model for the CoCoP approach.
This model is based on LLaMA2 and improves the coding capability of the base model.
Subsequently, we conducted a comparative analysis between the CodeLLaMA-based models and the baseline LLaMA2 model.
We report results for the 7B, 13B, and 34B code models.

LLaMA2 and CodeLLaMA models are available in 7B and 13B sizes, allowing for a comparative analysis of CoCoP's impact on their performance.
\Cref{tab:Models size} presents the results of this evaluation.
For the 7B size, CodeLLaMA with CoCoP outperformed LLaMA2 when employing CoCoP and few-shot learning techniques across all four datasets.
Notably, using CoCoP with CodeLLaMA significantly enhanced accuracy, exceeding 20\% for SST2 and SNLI datasets, compared to using LLaMA2 with CoCoP and few-shot learning methods.
In the 13B model size, CodeLLaMA with CoCoP outperformed LLaMA2 with CoCoP and few-shot learning in SST2, CoLA, and SLI.
However, LLaMA2 with few-shot learning showed slightly better performance with a 0.8\% improvement in MRPC.
These results show that using CoCoP with a code model has the potential to improve performance for text classification tasks significantly.
This method also works well with small code models.
Note that each experiment was conducted ten times with a different set of examples for both the few-shot and CoCoP methods, and the standard deviations were reported accordingly.
The results demonstrated that the CoCoP with code model exhibited lower standard deviations than the other approaches.
This finding suggests that CoCoP is more robust to variations in example contexts and exhibits greater reliability.

\subsubsection{Few-shot learning vs. CoCoP in code models}
This subsection evaluates the difference between CoCoP and few-shot learning methods in CodeLLaMA as a code model.
\Cref{tab:Models size} shows the results of these evaluations.
CoCoP performed better in all four datasets in all model sizes, including 7B, 13B, and 34B.
These results mean that performance improvement through CoCoP with CodeLLaMA is independent of this model's capability to do classification or learning through demonstrations and comes from the CoCoP method.
Another surprising result from this experiment is that few-shot learning performance decreases in CodeLLaMA by increasing model size.
\subsubsection{Models size}
In \Cref{sec:few-shot learning vs. CoCoP}, we observed that employing the CoCoP method shows better accuracy than the few-shot technique when using LLaMA2 models of the same size.
Additionally, \Cref{sec:coders vs. base models} highlighted that utilizing the CodeLLaMA-Instruct model shows better accuracy results than the LLaMA2 model when using the CoCoP method, despite both being in the same model size.
This subsection underscores the assessment that employing the CoCoP method with a smaller code model can yield better results than larger LLaMA2 models without CoCoP.
As the model size grows, the inference cost increases.
Therefore, it is essential to attain improved results by using smaller models.

As can be seen in \Cref{tab:Models size}, using the CodeLLaMA-34B-Instruct with the CoCoP method yielded better results than utilizing the LLaMA2-70B-chat model with the few-shot method across SST2, CoLA, and MRPC datasets.
The results obtained using the CodeLLaMA-34B-Instruct in the SST2 and CoLA datasets exhibited about 10\% improvement in classification accuracy.
Additionally, it demonstrated comparable performance in the SNLI dataset.
This suggests that the CoCoP method with the CodeLLaMA model can achieve better or comparable results than the traditional few-shot method despite using a twice-as-small model.

Another unexpected finding is that when employing CodeLLaMA-7B-Instruct and CodeLLaMA-13B-Instruct with the CoCoP method, we observed better results in the SST2 and CoLA datasets than utilizing LLaMA2-70B-chat with a few-shot technique.
Moreover, these models exhibit results comparable to those of the LLaMA2-70B-chat with a few-shot method in the MRPC and SNLI datasets with only about 4\% difference accuracy.
These observations suggest that achieving high performance in classification tasks is feasible with smaller LLMs.
These results suggest a trend towards developing prompt techniques for smaller language models that can maintain performance levels comparable to larger models.
\subsubsection{Providing additional information in the prompt}
Additional information can be added to the generated prompt based on CoCoP's prompt generation in code format.
For instance, CoCoP can show all possible labels for classification with a list of strings in the code format.
We assess the impact of adding additional information to generate incomplete-code in CoCoP.

\begin{figure}[ht]
    \centering
    \includegraphics[width=\columnwidth]{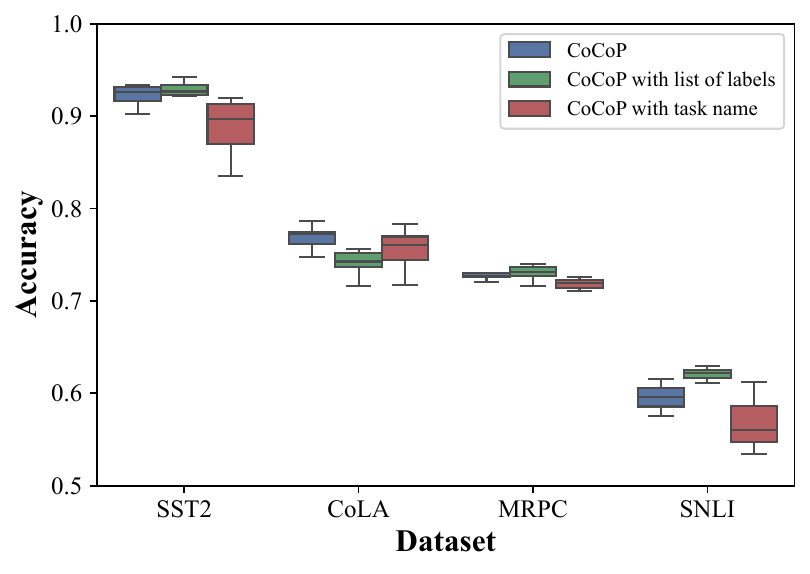}
	\caption{The accuracy metric evaluated the comparison of the standard CoCoP method with the adaptation of including task names or a list of all potential labels while generating incomplete code across all available datasets.}
	\label{fig:CodeInstruction}
\end{figure}

\begin{figure*}[ht]
    \centering
	\includegraphics[trim=0cm 2cm 0cm 0cm,width=0.85\textwidth]{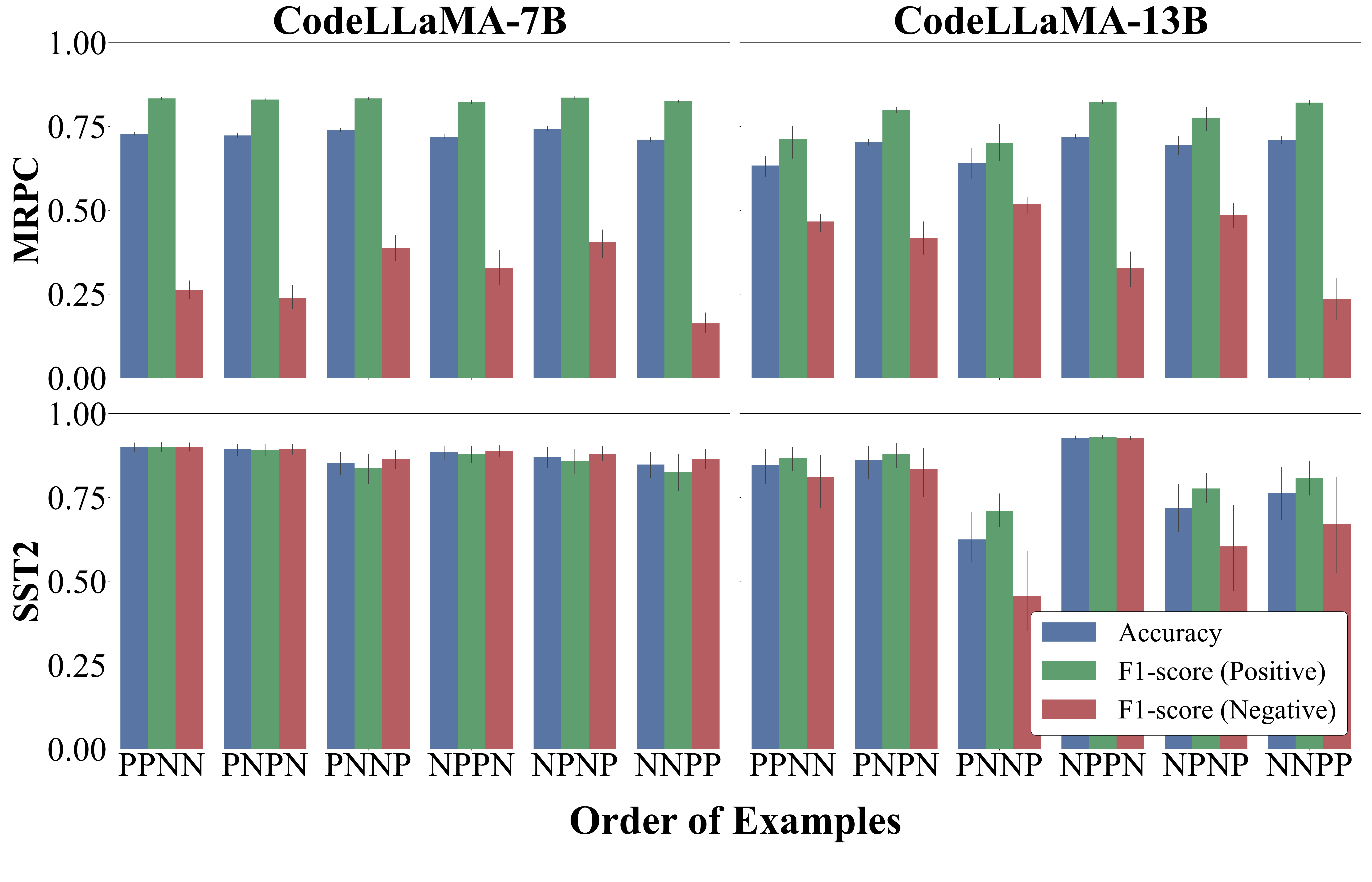}
	\caption{The impact of examples' order in the CoCoP method on the MRPC and SST2 datasets.}
	\label{fig:shot order - SST2}
\end{figure*}

To guide the LLM in determining the label, we consider two types of additional information: the names of all possible labels and the task's name.
In the first scenario, the Incomplete-Code Generator adds a line at the beginning of the prompt. 
This line shows a variable assignment that assigns a list of strings to a variable.
Each string represents a possible label.
For the second scenario, the Incomplete-Code Generator module adds the name of each task at the beginning of the prompt as a code comment.
\Cref{fig:CodeInstruction} shows the results.

Results showed that adding a list of labels slightly improved SST2, MRPC, and SNLI, but accuracy decreased in CoLA.
On the other hand, adding a task's name decreased accuracy in all datasets.
These results show that adding extra information has the potential to improve CoCoP performance, but the type of extra information is critical.

\subsection{Ablation Study}
Our proposed method incorporates certain characteristics which may influence its performance.
To assess the impact of these characteristics, we conducted a series of experiments using CodeLLaMA-7B and CodeLLaMA-13B models on the MRPC and SST2 datasets.
The choice of these code models was motivated by the demonstrated efficacy of the CoCoP method in enhancing their performance.
Specifically, we selected SST2 as a dataset where CoCoP has shown significant improvements, while MRPC represents a task where CoCoP's performance gains are not as substantial, albeit still present.
Our analysis focused on two key characteristics: the structure of the incomplete code and the type of the examples used in the incomplete code prompt.
We conducted separate experiments to isolate and evaluate the impact of each of these characteristics.

\subsubsection{CoCoP prompt structure}
\begin{figure*}[ht]
    \centering
	\includegraphics[trim=0cm 26cm 0cm 0cm,width=0.95\textwidth]{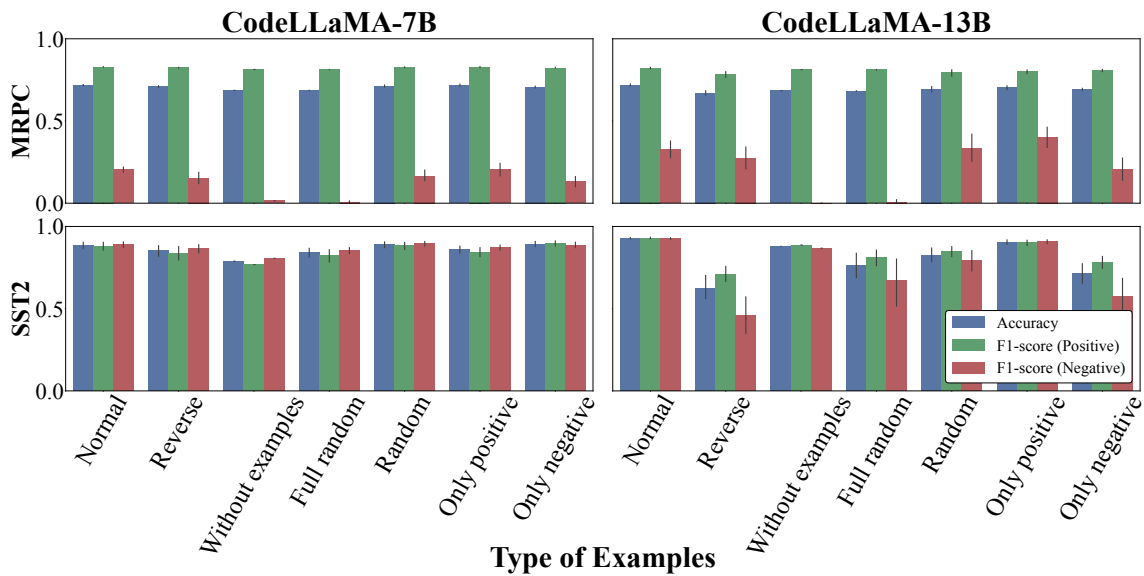}
	\caption{Effect of example type variation in CoCoP models on the MRPC and SST2 dataset.}
	\label{fig:shots type - SST2}
\end{figure*}
The CoCoP method uses the concept of few-shot learning by using demonstrations.
These demonstrations can be arranged in different orders.
To investigate this, we conducted experiments using two positive examples and two negative examples with varying orders.
\Cref{fig:shot order - SST2} illustrates the results for both datasets with two model sizes.
These results showed that CodeLLaMA-7B is less sensitive to example orders than CodeLLaMA-13B.
Specifically, in MRPC, the order of examples has a greater influence on the F1-score of the negative class than accuracy and the positive class.
Hence, this ordering can significantly affect the performance of the negative class.

\subsubsection{CoCoP examples type}
The type of demonstrations can impact performance in few-shot learning~\citep{min2022rethinking}. Therefore, we conducted experiments to evaluate the impact of different types of examples in the CoCoP method.
By Keeping the incomplete-code structure fixed, we varied the type of examples used.
This entails maintaining the order of examples based on their labels while allowing for changes in the sentences themselves.
Seven different types of example sentences were employed:
\begin{itemize}
    \setlength\itemsep{-5pt}
    \item \textbf{Normal:} Basic version.
    \item \textbf{Reverse:} Replaces negative class with a positive sentence and positive class with a negative sentence.
    \item \textbf{Without examples:} Only shows labels without any accompanying sentences.
    \item \textbf{Full random:} Replaces sentences with random ones from other datasets.
    \item \textbf{Random:} Replaces sentences with random ones from its own dataset.
    \item \textbf{Only positive:} Replaces all sentences with positive ones.
    \item \textbf{Only negative:} Replaces all sentences with negative ones.

\end{itemize}

\Cref{fig:shots type - SST2} shows the results for MRPC and SST2.
In SST2, CodeLLaMA-7B was less sensitive to the type of examples than CodeLLaMA-13B.
However, in MRPC, experiments with ``Without Examples" and ``Full random" types indicated that the F1-score for the negative class is close to zero.
This shows that the model lacks sufficient context about the negative class when no examples are provided, and the type of examples becomes crucial.
Notably, employing examples from its dataset yields improved results.

\section{Related Work}
\label{sec: related work}
\subsection{Text classification by LLMs}
The introduction of the transformer model~\citep{vaswani2017attention} has led to the development of numerous models based on this architecture.
One of the most influential models is BERT, which is pre-trained on a vast amounts of data and can be fine-tuned for downstream tasks, such as classification~\citep{kenton2019bert}.
Following the BERT model, the advent of LMs like GPT-2~\citep{GPT-2_few-shot_learner} and T5~\citep{2020t5} has further expanded the capabilities of transformer-based models in various downstream tasks.

Two primary approaches have emerged for leveraging these models in classification tasks: fine-tuning and prompt engineering~\citep{kaddour2023challenges}.
Fine-tuning involves further training the pre-trained LLM on a specific task and dataset, allowing the model to adapt its parameters to the target domain.
Prompt engineering, on the other hand, involves carefully crafting natural language prompts to elicit desired responses from the frozen LLM without further fine-tuning.
Given the large parameter sizes of LLMs such as GPT-4~\citep{achiam2023gpt4}, Gemini~\citep{team2023gemini}, and LLaMA2~\citep{llama2}, we employ the concept of prompt engineering.
This approach enables seamless adaptation to various LLMs without necessitating additional resources for fine-tuning. 
\subsection{LLMs for code (code models)}
Some LLMs are fine-tuned on specific tasks to enhance their performance in specialized domains.
One of the most significant tasks for LLMs is code-related tasks, such as code generation and code completion.
For instance, Codex by OpenAI~\citep{chen2021Codex} and PalmCoder by Google~\citep{chowdhery2023palm} are examples of proprietary, closed-source code models.
On the other hand, open-source code models, such as CodeLLaMA~\citep{roziere2023codeLlama}, CodeGemma~\citep{codegemma_2024}, WazirCoder~\citep{luo2023wizardcoder}, and InCoder~\citep{fried2022incoder}, are also available.
We call these models code models.
While code models are primarily employed for code-related tasks, the CoCoP method leverages their capabilities beyond the conventional scope of code processing.
Specifically, we harness the models' prowess in code-related tasks to perform a classification task, thereby extending their utility to a broader range of applications.

\subsection{Performing tasks with LLMs through code format}
Several previous works have leveraged the capabilities of LLMs in code-related tasks to tackle problems beyond the coding domain.
For instance, Code as Policy (CaP)~\citep{liang2023codeAsPolices} employed LLMs to generate code policies for robots.
Another example is TidyBot~\citep{wu2023tidybot}, which utilized LLMs' code completion capabilities to perform robotic household cleanup tasks. 
Both approaches presented their problems as incomplete-code to the LLM, which then completed the code to accomplish the desired tasks.

Additionally, there are works like Chain of Code~\citep{li2023CoC} and Program of Thought~\citep{chen2023PoT} that use the Chain-of-Thought (CoT)~\citep{wei2022CoT} concept and apply it to LLMs' code-related capabilities to enhance the models' reasoning performance.
The CoCoP leverages the capabilities of LLMs in code-related tasks, similar to previous works .
However, our method applies this capability to text classification problem.
%
\section{Conclusion}
\label{sec : conclusion}
Based on LLMs' excellent performance in code-related tasks, we proposed the Code Completion Prompt (CoCoP) method.
This method transforms classification tasks into code completion task and leverages few-shot learning.
Our evaluation of this method across many classification datasets demonstrated its efficacy in enhancing classification results.
For instance, CoCoP can improve the accuracy of the LLaMA2-70B-chat model in SST2 and CoLA benchmarks about 10\%.

Furthermore, we employed pre-trained and fine-tuned LLMs on code-related datasets (code models), and these models exhibited superior performance compared to basic models when using the CoCoP method.
CodeLLaMA-34B-Instruct with CoCoP outperformed LLaMA2-70B-chat with about 10\% accuracy improvement in SST2 and CoLA and 2\% improvement in MRPC.
Remarkably, smaller models like CodeLLaMA-7B-Instruct, when combined with the CoCoP method, attained comparable results to that of the LLaMA2-70B-chat model using the few-shot technique. 
For instance, CodeLLaMA-7B-Instruct with CoCoP outperformed LLaMA2-70B-chat in SST2 and CoLA and also showed comparable accuracy in MRPC and SNLI only with less than 4\% different accuracy.
These results showed better or comparable accuracy with one-tenth the model size compared to basic models
%
\section{Limitations and Future Work}
\label{sec: lim&f }
This study uses the capability of LLMs for code completion in classification tasks.
Much research work can be done based on this research.
While we precisely assess the CoCoP method in some important text classification datasets, it is conceivable that the proposed idea can be applied to a broader spectrum of text classification tasks.
Furthermore, several other classification tasks could be reconfigured as text classification problems and addressed using this method.

Based on the success of the CoCoP method in the text classification tasks, additional research is required to evaluate the effectiveness of this method in generation tasks.
Future work can focus on transforming the reasoning and generation tasks into a code completion problem and solving them using the CoCoP method.

This study has demonstrated that LLMs designed for specific tasks, such as code generation, can leverage their capabilities in domains beyond their original intended purpose.
The findings suggest that future research endeavors could explore the potential of repurposing domain-specific LLMs for novel applications.
By adapting and fine-tuning these models on relevant datasets, it may be possible to harness their strengths and transfer their knowledge to new domains, thereby expanding the frontiers of their utility.

\section*{Acknowledgement}
Authors used OpenAI~ChatGPT-3.5 and Claude-3.5~Sonnet to post-edit the manuscript.

\bibliography{coling_latex}

\end{document}